\documentclass[runningheads]{llncs}
\usepackage{multirow}
\usepackage{url}
\usepackage{adjustbox}

\usepackage{cite}
\usepackage{amsmath,amssymb,amsfonts}
\usepackage{algorithmic}
\usepackage{algorithm}
\usepackage{xcolor}
\usepackage{graphicx}
\usepackage{textcomp}
\usepackage{ragged2e}
\usepackage{marvosym}

\begin{document}
\mainmatter 
\setcounter{secnumdepth}{4}
\title{Prompting GPT-3.5 for Text-to-SQL with De-semanticization and Skeleton Retrieval}

\titlerunning{Prompting GPT-3.5 for Text-to-SQL} 
%

\author{Chunxi Guo, Zhiliang Tian\textsuperscript{ (\Letter)}, Jintao Tang\textsuperscript{ (\Letter)}, Pancheng Wang, Zhihua Wen, \\Kang Yang and Ting Wang\textsuperscript{ (\Letter)}
}
\authorrunning{Guo et al.} 
\institute{College of Computer, National University of Defense Technology, Changsha, China\\
\email{\{chunxi, tianzhiliang, tangjintao, wangpancheng13, zhwen, yangkang, tingwang\}@nudt.edu.cn}\\
}
\maketitle 

\begin{abstract}

Text-to-SQL is a task that converts a natural language question into a structured query language (SQL) to retrieve information from a database. Large language models (LLMs) work well in natural language generation tasks, but they are not specifically pre-trained to understand the syntax and semantics of SQL commands.
In this paper, we propose an LLM-based framework for Text-to-SQL which retrieves helpful demonstration examples to prompt LLMs. 
However, questions with different database schemes can vary widely, even if the intentions behind them are similar and the corresponding SQL queries exhibit similarities. Consequently, it becomes crucial to identify the appropriate SQL demonstrations that align with our requirements.
We design a de-semanticization mechanism that extracts question skeletons, allowing us to retrieve similar examples based on their structural similarity.
We also model the relationships between question tokens and database schema items (i.e., tables and columns) to filter out scheme-related information. 
Our framework adapts the range of the database schema in prompts to balance length and valuable information. A fallback mechanism allows for a more detailed schema to be provided if the generated SQL query fails.
Ours outperforms state-of-the-art models and demonstrates strong generalization ability on three cross-domain Text-to-SQL benchmarks.
\end{abstract}

\section{Introduction}
Text-to-SQL tasks aim to transform natural language questions (NLQ) into structured query language (SQL), enabling users without expertise in database querying to retrieve information from a database~\cite{ratsql,Cai_Xu_Zhang_Yang_Li_Liang_2018_Encoder-Decoder}.
Considering that databases are used in various scenarios involving different domains (e.g., education, financial systems), researchers have adapted encoder-decoder architecture~\cite{li2023resdsql, Graphix-T5}, which eliminates the need for domain-specific knowledge through end-to-end training. To train the model, these approaches require diverse and extensive training data, which can be prohibitively expensive~\cite{zhao2023survey}.



Large pre-trained language models (LLMs) (e.g., GPT-3~\cite{GPT3} and Codex~\cite{Codex}) encompass more extensive data and parameters than traditional pre-trained language models (e.g., BERT \cite{bert}, RoBERTa \cite{RoBERTa}, BART \cite{BART} and T5 \cite{t5}) and exhibit superior performance on a variety of tasks, including Text-to-SQL.
Rajkumar et al. \cite{EvaluatingLLM} and Liu et al. \cite{Evaluatingchatgpt} evaluate LLMs' performance in Text-to-SQL in zero- and few-shot settings. 
Cheng et al. \cite{Binding} present a neural-symbolic framework that maps the input to a program, which incorporates symbolic components into LLMs.
However, many studies found that LLMs perform worse than traditional non-LLM-based approaches in Text-to-SQL \cite{picard,rasat,li2023resdsql}. As the existing LLMs are not designed for understanding the syntax and semantics of SQL commands, it is challenging for them to accurately generate complex SQL commands (e.g. \textit{SELECT}, \textit{WHERE}, \textit{AVG}, \textit{DESC}). 
To accurately map out these SQL commands, it is essential to distinguish the question intention.
Intention in Text-to-SQL tasks refers to the collection of query-related specifications and directives that encompass the desired scope, criteria, and actions to be performed on a database. This concept encompasses various desired result set attributes, including data volume, sorting sequence, and filtering prerequisites. For example, the skeleton corresponds to the question \textit{``What are the names of the singers who are not French?''} is \textit{``What are the [MASK] of the [MASK] who are not [MASK]?''}, whose intention is to query a term with a conditional constraint.
\vspace{-3mm}
\begin{figure}[H]
\centering
\includegraphics[width=1.0\textwidth]{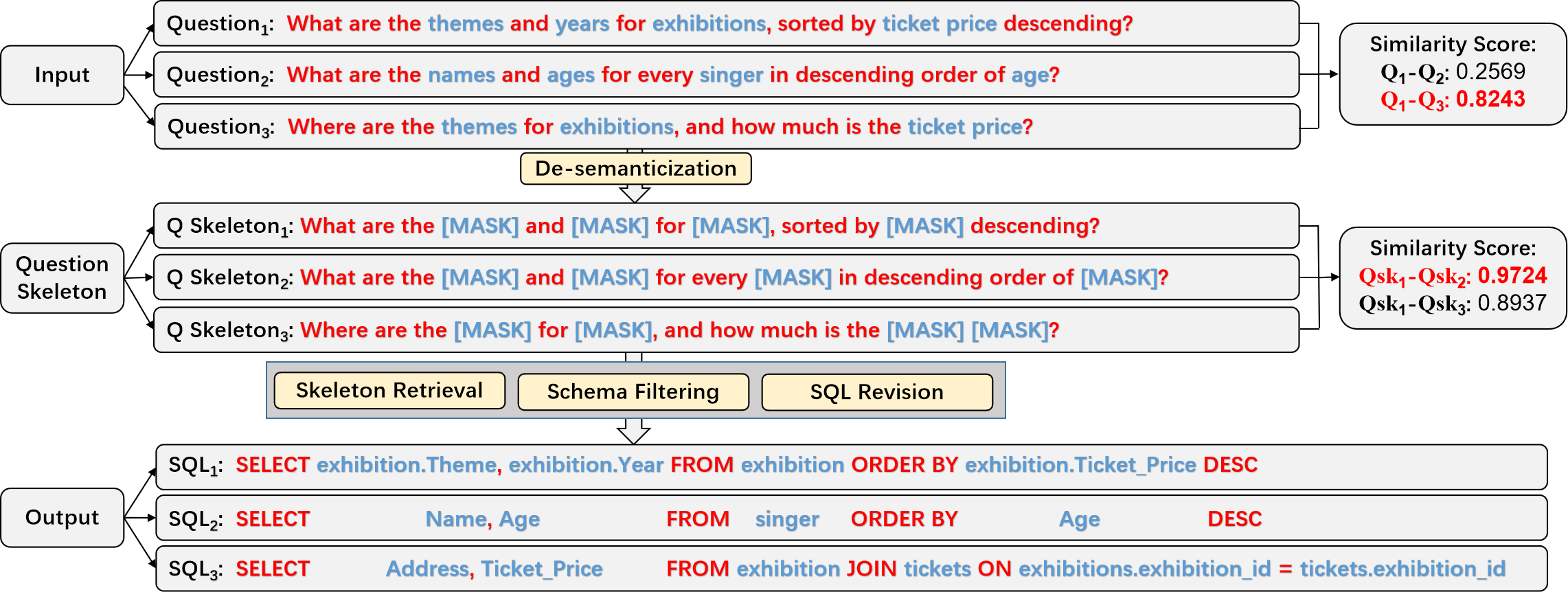}
\caption{Comparison of three examples (i.e. question, question skeleton, SQL).  The first example is similar to the second in terms of question intention (sorting), and to the third in terms of vocabulary of the questions. Note that the intention of the third question is to get two attribute items and there may be a table join.
We aim to obtain SQL queries with the same commands (i.e. \textit{ORDER BY}, \textit{DESC}) in the prompt. 
Compared with the full question similarity score, the question skeleton increases the absolute value as well as the relative ranking. 
} \label{fig:intro}
\end{figure}
\vspace{-3mm}

LLMs are proven to fast adapt to new paradigms with few-shot examples \cite{lee2022does, Su_Kasai_2022, Rubin_Herzig_Berant}. We argue that LLMs probably quickly learn to follow some demonstration examples of SQL generation, even if the SQL generation involves multiple SQL commands and nested clauses.

In this paper, we propose an LLM-based Text-to-SQL framework that retrieves a few demonstration examples to prompt the LLM according to the skeleton of the input question. 
Notice that questions with different database schemes may be distinct since questions contain much scheme-related information (i.e. the blue texts in Fig.~\ref{fig:intro}'s upper side), even if they have similar intention and SQL queries. It can be difficult for the model to retrieve helpful examples.
To solve this issue, we design a de-semanticization mechanism to extract skeletons of questions. We retrieve similar SQL demonstrations, which share similar question skeletons with the input question. 
Besides, during de-semanticization, we model the relationships between question tokens and scheme items to filter out the scheme items related to the question tokens. In this way, we achieve schema linking concerning the question-scheme relevance. 
Finally, we adaptively control the range of the database schema in prompts to balance length and valuable information. Through a fallback mechanism, the model receives a more detailed schema if the generated SQL query needs revision.
Our framework excels compared to commercial Text-to-SQL engines like AI2sql\footnote{https://app.ai2sql.io} by emphasizing transparent, flexible algorithms for better performance and deeper insights into the conversion process, whereas comparisons to large models like GPT-4 might not be entirely fair due to data volume discrepancies.

Our contributions are as follows:
(1) We propose an LLM-based framework for Text-to-SQL tasks that retrieves similar examples to augment prompts for LLMs.
(2) We design a de-semanticization mechanism that effectively removes scheme-related information for retrieving texts with similar skeletons. 
(3) Our method surpasses the SOTA models and exhibits strong generalization.

\section{Related Work}
The evolution of SQL generation techniques showcases a progression from encoder-decoder architectures \cite{Cai_Xu_Zhang_Yang_Li_Liang_2018_Encoder-Decoder} to LLM-based solutions.

\textbf{Encoder-based SQL Generation.} 
Guo et al. \cite{irnet} introduced IRNET, utilizing attention-based Bi-LSTM to encode and an intermediate representation-based decoder for SQL prediction. Afterwards, graph-based encoders were integrated to enhance input representations \cite{Bogin_Berant_Gardner_2019, Chen_Chen_Zhao_Cao_Xu_Zhu_Yu_2021_ShadowGNN}.
Works such as RATSQL \cite{ratsql}, LGESQL \cite{LGESQL}, R2SQL \cite{Hui_Geng_Ren_Li_Li_Sun_Huang_Si_Zhu_Zhu_2021}, SDSQL \cite{li2023resdsql}, S2SQL \cite{hui-etal-2022-s2sql}, and STAR \cite{STAR_Cai_2022} focused on refining structural reasoning by explicitly modeling relationships between schemas and questions. 
Notably, GRAPHIX-T5 \cite{Graphix-T5} overcame prior limitations by incorporating graph representation learning into the encoder. Simultaneously, RASAT \cite{rasat} augmented T5 with structural insights by introducing edge embeddings into multi-head self-attention.

\textbf{Decoder-based SQL Generation.} 
We categorize the methods into four distinct groups:
Sequence-based methods like BRIDGE \cite{lin2020bridging} and PICARD \cite{picard} directly translate natural language queries into SQL queries token by token. 
Template-based methods, represented by X-SQL \cite{X-SQL} and HydraNet \cite{hybrid}, utilize predefined templates to guide SQL generation, ensuring structural coherence. 
Stage-based methods, exemplified by GAZP \cite{GAZP} and RYANSQL \cite{choi-etal-2021-ryansql}, involve establishing a coarse-grained SQL framework, subsequently employing slot-filling methodologies to complete missing details within the framework. Lastly, hierarchical-based methods, such as IRNet \cite{irnet} and RAT-SQL \cite{ratsql}, adopt a hierarchical approach to tackle NLQ-to-SQL translation.

\textbf{LLM-based SQL Generation.} 
Recently, LLM-based models are now prominent options for this task.
Unlike full-data fine-tuned models, LLM-based models can achieve good performance with just a few unsupervised in-context exemplar annotations \cite{zhao2023survey}. Yu et al. \cite{yu2021similar} introduced a method to classify and cluster SQL queries based on question characteristics.
Inspired by some retrieval-related research~\cite{tian2019learning,song2022retrieval,wen2023grace}, we retrieve SQL examples with the same intention as a demonstration, thereby enhancing the comprehension of the diverse operators and their respective applications. Our approach improves the LLM's performance to generate valid and accurate SQL queries.

\section{Methodology}
\begin{figure}[t] \centering
 \includegraphics[width=1.0\textwidth]{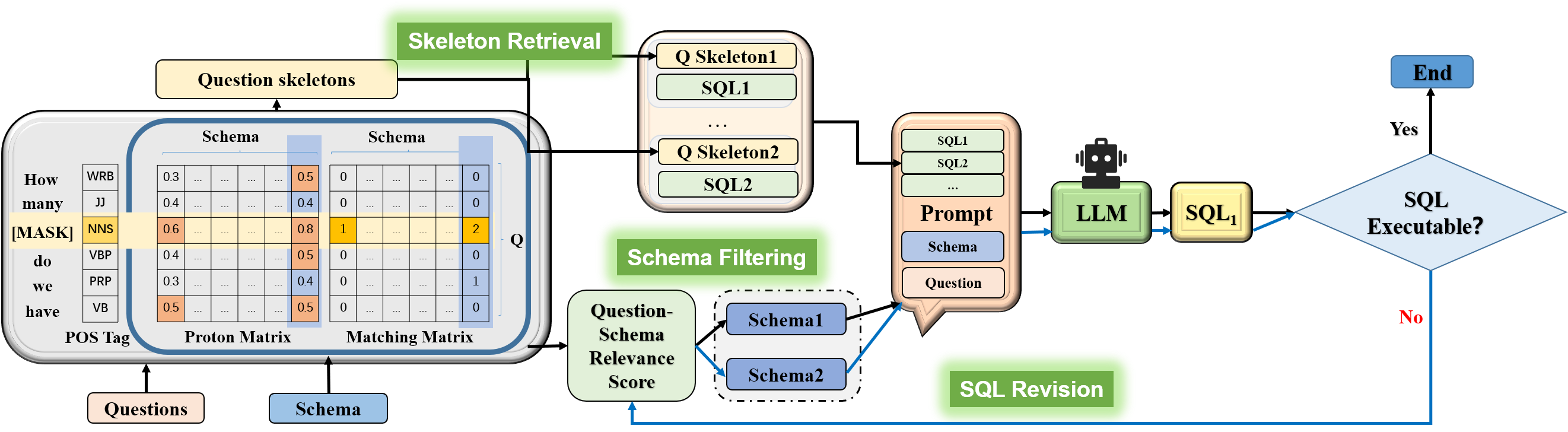}
 \caption{The overview of our framework. The grey box on the left shows the details of the de-semantization process. The rest is the process of prompt construction. We revise the SQL following the blue line.} \label{fig:model}
\end{figure}

Our framework consists of two modules as shown in Fig.~\ref{fig:model}: 
(1) \textbf{Question De-semanticization} (Sec.~\ref{De-semanticization}) removes tokens that are semantically related to the domain and preserves the question skeletons, which represent the question's intentions. 
(2) \textbf{LLM-Based Adjustable Prompting} (Sec.~\ref{Prompt Construction}) involves using the SQL demonstrations with the same intention and corresponding database schema to create prompts that guide the LLM in generating SQL queries.



\label{Representation}

Given a natural language question ${Q}$ and the database schema ${S}=\langle{T}, {C}\rangle$, the goal of Text-to-SQL tasks is to generate the corresponding SQL $P$. Here the question ${Q}=\left (q_1, q_2, \ldots, q_{|{Q}|}\right)$ is a sequence of words. The database schema consists of tables ${T}=\left (t_1, t_2, \cdots, t_{|{T}|}\right)$, and columns ${C}=\left (c_1, c_2, \cdots, c_{|{C}|}\right)$. 


\subsection{Question De-semanticization}
\label{De-semanticization}
We remove question tokens that are semantically related to the database schema (i.e., table, column) and obtain the question skeletons, which represent the question's intentions.
In this way, from a matching perspective, by eliminating the schema-related information from the question, we can better match examples with similar question intent when retrieving.

There are two steps involved in this process:
the first step is schema linking, which finds the question tokens related to the schema. Meanwhile, we obtain the relevance of the question tokens to the schema.
As a second step, we mask the tokens and obtain the skeletons of the questions.
In the first step, we cannot determine how relevant the tokens are to the schema, so we design schema-related detection (Sec.~\ref{Schema-Related Detection}), token matching (Sec.~\ref{Token Matching}) and part-of-speech tagging (Sec.~\ref{pos}) strategies to find relevant tokens.
\subsubsection{Schema-Related Detection.}
\label{Schema-Related Detection}
We use a masking technique to identify correlations between question words and their corresponding database schema.
Concretely, we calculate the similarity between question tokens and schema items. There are three steps:

\textbf{(1) Masking and Concatenation.} We concatenate a question ${Q}$ and schema (i.e., table ${T}$, column ${C}$) into one long sequence. We also mask each token of the question in the sequence to generate a series of sequences.
Formally, we obtain a series of sequences as follows, where [CLS] and [SEP] denote classification and sentence separation, respectively:
\begin{small}
$$
[CLS] q_1  q_2\cdots q_{|{Q}|} [SEP] [CLS] t_1 \cdots t_{|{T}|}[SEP] [CLS] c_1\cdots, c_{|{C}|} [CLS]
$$
$$
[CLS] [MASK] q_2 \cdots q_{|{Q}|} [SEP] [CLS] t_1 \cdots, t_{|{T}|} [SEP] [CLS]c_1\cdots, c_{|{C}|} [CLS]
$$
$$
\cdots
$$
\end{small}

Later, we put all sequences into a pre-trained language model to obtain deep contextualized representations.
We denote ${h}_j^s$ as the representation of schema item $s_j$ and ${h}_{j \backslash q_i}^s$ as the representation if the question token $q_i$ is masked out.

\textbf{(2) Representation Transformation.} 
We utilize the standard Poincaré ball~\cite{Hyperbolic}—a unique model of hyperbolic spaces to project the representations.
We get the hyperbolic representations, denoted as $\tilde{h}$, using the following method:
\begin{equation}
\label{h}
\tilde{\mathrm{h}}=g_0(\mathrm{~h})=\tanh (\|\mathrm{h}\|) \frac{\mathrm{h}}{\|\mathrm{h}\|}.
\end{equation}

The hyperbolic space provides a suitable geometry for modeling the hierarchy~\cite{Hyperbolic}. The hyperbolic space has a property called a negative curve which allows for a more efficient representation of the hierarchy, enabling our method to capture long-term dependencies and the overall sentence structure~\cite{Probing_BERT}. Furthermore, the hyperbolic space has a greater power to represent than the Euclidean space low-dimensional space, thus allowing for a more efficient representation of sentences with semantic hierarchy.

\textbf{(3) Correlation Measurement.} 
We measure the correlation between the question token $q_i$ and the schema item $s_j$ in the hyperbolic space.
Concretely, we elicit the correlation between question tokens and schema items from a pre-trained language model based on the Poincaré distance matrix~\cite{Probing_BERT}.

By computing $d_p$ on each pair of tokens $\left (q_i, s_j\right)$, we get the Proton matrix $\mathrm{D_p} \in \mathbb{R}^{|Q|\times|S|}$ as follows:
\begin{equation}
\label{d}
\mathrm{D_p}_{i,j} = 2 \tanh ^{-1}\left (\left\|-\tilde{h}_{j \backslash q_i}^s \oplus \tilde{h}_j^s\right\|\right),
\end{equation}

where $\oplus$ is the Möbius addition~\cite{Hyperbolic}, $\tilde{h}_j^s$ represents the embedding of the schema item $s_j$, and $\tilde{h}_{j \backslash q_i}^s$ represents the embedding if the question token $q_i$ is masked out. Both $\tilde{h}_j^s$ and $\tilde{h}_{j \backslash q_i}^s$ are hyperbolic representation defined in Eq. (\ref{h}).


We argue that the sequence concatenating all tables and columns represents the domain knowledge of the example, which consists of the vocabulary of the domain/scenario. We mask each token in the question sequence to detect whether it is relevant to the domain.
This measure works by masking the token and checking if it causes a significant shift in the vector representation of the entire sequence.  If the shift is beyond a pre-defined threshold, we consider the masked token to be important for the sequence.  This means that the token is strongly correlated with the meaning expressed by the sequence.





\subsubsection{Token Matching.}
\label{Token Matching}
To discover the question tokens closely related to the scheme, we match each question token and each schema item (i.e., table names, column names, and values) with two kinds of explicit information in the input: name-based and value-based matching. 


Name-based matching identifies direct lexical matches between each question token and each schema item.
If a sub-sequence of the question sequence $q_{i...j}$ matches the schema names $s_j$, score one point for matching similarity.
While value-based matching detects possible value correspondences within the query. If the question word $q_i$ is equal to specific values $v_j$ in the database, where $v_j$ represents the set of values in the $j^{th}$ column of the corresponding table or column, score one point for the corresponding matching similarity. 
We define matrix $M_m \in \mathbb{R}^{|Q|\times|S|}$ to represent the question-schema matching similarity: 

\begin{equation}\label{m}
M_{m_{i,j}} =
\begin{cases}
2, & q_{i...j}  \subseteq  s_j \wedge q_i = v_j \\
1, & q_{i...j}  \subseteq  s_j \vee q_i = v_j \\
0, & \text{otherwise}
\end{cases}.
\end{equation}


So far, coupled with the previously calculated Proton matrix $\mathrm {D_p}$, we get the question-schema relevance score, which measures the probability that the schema items will be used to compose the SQL query.
We define the relevance score matrix $R \in \mathbb{R}^{|Q|\times|S|}$ as follows:
\begin{equation} \label{R}
R = \mathrm{D_p} + \beta \cdot M_m ,
\end{equation}
where $\beta$ determines the relative influence of these two strategies.

\subsubsection{Part-of-Speech Tagging.}
\label{pos}
We perform part-of-speech (POS) tagging on questions to improve the recognition of the question skeleton.

Formally, we tag the question $Q$ with POS analysis to obtain a set of POS tags ${t_1, t_2, \dots, t_n}$, where $t_i$ is the POS tag of a token $q_i$. Then we generate the lexical matrix $P \in \mathbb{R}^{|Q|}$ for each token $q_i$ based on its POS tag $t_i$ as follows:

(1) If $t_i$ is a noun or a number, then $P_i$ is assigned the value $\alpha$.

(2) Otherwise, $P_i$ is assigned the value 0.

POS information is crucial for constructing a question skeleton as it aids in comprehending the sentence's structure and the grammatical roles of the words.


Incorporating the three strategies mentioned above, we obtain a question relevance score ${Q_{\text {sco}}}_i$ for each question token $q_i$ using the following equation:
\begin{equation}
{Q_{\text {sco}}}_i = \frac{1}{2} \left ( \frac{1}{n} \sum_{j=1}^{|S|} R_{ij} + P_i \right), \quad \forall i \in {1, \dots, |Q|}.
\end{equation}
We generate the question skeleton based on $Q_{\text {sco}}$ and $\tau$, where $\tau$ is a hyperparameter that controls the minimum relevance score required for a token.\footnote{If $q_{\text {sco}}$ is below the threshold of $\tau$, we retain the original question token; otherwise, we replace it with the pre-defined [MASK] token.}

After calculating the $R$ (in Sec.~\ref{Token Matching}) and $P$, we remove domain-relevant tokens of the questions and generate the de-semanticized question skeletons, which allows for better retrieval of examples where the question intentions are more consistent and more applicable to the in-context demonstration.

\subsection{LLM-Based Adjustable Prompting}
\label{Prompt Construction}
To construct prompts for the LLM to generate new SQL queries, we utilize the SQL queries obtained from the question skeleton and the relevant database schema, which are filtered by the question-schema relevance score.
Then we revise the SQL queries via a fallback mechanism, which adjusts the schema range.
The prompt we designed consists of three parts as shown in Fig.~\ref{fig:model}.


\subsubsection{$K$NN-Based Skeleton Retrieval.} 
\label{retrieval}
We retrieve $k$-NN examples based on the new question skeleton.
We project de-semanticized question skeletons into a vector space and retrieve $k$-NN examples corresponding to the new question skeleton. We use cosine similarity to measure the text vectors.
The new question skeleton serves as the key for the retrieval process, and the returned value consists of the $k$-NN examples. 
Questions with the same intention can aid in generating SQL queries by sharing common structures and requiring comparable SQL queries to extract information from a database. Recognizing patterns and similarities between questions enables the language model to produce suitable SQL queries for a given inquiry.

\subsubsection{Schema-Relevance Filtering.} 
We utilize the filtered database schema items to generate SQL queries.
We get the schema with a scaled-down range by the relevance scores between the question and the schema.
We apply a threshold $\theta$ to filter out less relevant schema items based on the question-schema relevance score obtained in Sec.~\ref{De-semanticization}. Specifically, we only consider schema items with scores higher than the $\theta$.
with a scaled-down schema range, we prompt the LLM to generate the SQL queries.
Narrowing the schema range prevents irrelevant schema from interfering with the model's response to the current question.

\subsubsection{Fallback Revision.} 
\label{revision}
We propose a fallback mechanism to revise and regenerate SQL queries, in cases where the LLM outputs a message indicating SQL generation failure or when the generated SQL query cannot be executed successfully.
We first check if the generated SQL query is valid and can be executed on the database. If the query fails, we retrieve the complete database schema and use it to revise the SQL query. This ensures that the revised SQL query is valid and can be executed on the database. We then pass the revised SQL query to the LLM for further processing.
To avoid generating an infinite loop of fallbacks, we set a maximum number of fallback attempts. If the maximum number of fallback attempts is reached and the SQL query still cannot be generated, we terminate the process.

\section{Experiment}
\label{Experiment}
\subsection{Experimental Setup}

$\quad$

\textbf{{Datasets.}}
In our study, we perform experiments on three widely recognized benchmark datasets: 
(1) \textbf{Spider} \cite{spider} covers a diverse range of 138 domain databases, offering a large-scale evaluation platform. 
(2) \textbf{Spider-Syn} \cite{syn} is a modified version of Spider that introduces difficulty by replacing explicit question-schema alignments with synonymous phrasing. (3) \textbf{Spider-DK} \cite{dk} is an augmentation of Spider, incorporating artificial domain knowledge to further test model adaptability and comprehension.

\textbf{{Evaluation.}}
\textbf{Valid SQL (VA)} measures the percentage of SQL queries that are executed without any errors. \textbf{Execution accuracy (EX)} measures the accuracy of the execution results by comparing them with the standard SQL query. \textbf{Test-suite accuracy (TS)} \cite{ts} measures the effectiveness of the distilled test suite in achieving high code coverage for the database through execution, which can serve as a better proxy for semantic accuracy.
Note that we do not rely on the mainstream exact match accuracy metric (EM), as SQL queries that serve the same purpose may be expressed in different ways. EM is tailored to a limited style of the dataset and serves as an intermediate solution evaluation metric for Text-to-SQL tasks.


\textbf{{Baselines.}}
Full-data fine-tuned models: \textbf{PICARD} \cite{picard} employs incremental parsing to constrict auto-regressive decoders in language models;
\textbf{RASAT} \cite{rasat} enhances transformer models with relation-aware self-attention, combined with constrained auto-regressive decoders; and
\textbf{RESDSQL} \cite{li2023resdsql} introduces a novel framework featuring ranking-enhanced encoding and skeleton-aware decoding.
LLM-based models:
We select the \textbf{ChatGPT}~\cite{Evaluatingchatgpt} and \textbf{Codex}~\cite{EvaluatingLLM} baseline models for our study, as they are currently the top performers in evaluating Text-to-SQL capability using LLMs. Furthermore, our approach utilizes the latest \textbf{GPT-3.5} model, text-davinci-003. We all use the best performing with prompt engineering methodologies adapted to our respective models.


\textbf{{Experimental Setting.}}
We use FAISS \cite{FAISS} to store and retrieve question skeletons.
For hyperparameter settings, we assign $k$=8, $\alpha$=0.9, $\beta$=0.5, $\tau$=0.6, and $\theta$=0.4.
Our approach expands the database content beyond the limitations of the Text-to-SQL prompt used in the OpenAI demo website\footnote{https://platform.openai.com/examples/default-sqltranslate}, which only contains table and column names. We re-formatted the prompt to achieve better results, though it differs from the format used in the official data training.

\subsection{Main Results}

\begin{table*}[]
\centering
\renewcommand\arraystretch{1}
\tabcolsep=1.5cm
\small
\caption{Comparison of the performance of our model and others on three datasets. \protect\footnotemark[4]}
\setlength{\tabcolsep}{1mm}{
\begin{tabular}{cc|ccc|ccc|ccc}
\hline
\multicolumn{2}{c|}{\multirow{2}{*}{\textbf{Models}\textbackslash{}\textbf{Datasets}}} & \multicolumn{3}{c|}{\textbf{SPIDER}} & \multicolumn{3}{c|}{\textbf{SPIDER-SYN}} & \multicolumn{3}{c}{\textbf{SPIDER-DK}} \\ \cline{3-11} 
\multicolumn{2}{c|}{} & VA & EX & TS & VA & EX & TS & VA & EX & TS \\ \hline
\multicolumn{1}{c|}{\multirow{3}{*}{\begin{tabular}[c]{@{}c@{}}\textbf{Full-data}\\ \textbf{Fine-tuned}\\ \textbf{Models}\end{tabular}}} & PICARD & 98.4 & 79.3 & 69.4 & 98.2 & 69.8 & 61.8 & 97.8 & 62.5 & - \\ 
\multicolumn{1}{c|}{} & RASAT & 98.8 & 80.5 & 70.3 & 98.3 & 70.7 & 62.4 & 98.5 & 63.9 & - \\ 
\multicolumn{1}{c|}{} & RESDSQL-3B & \textbf{99.1} & 84.1 & 73.5 & 98.8 & 76.9 & 66.8 & 98.8 & 66.0 & - \\ \hline
\multicolumn{1}{c|}{\multirow{4}{*}{\begin{tabular}[c]{@{}c@{}}\textbf{LLM-based}\\ \textbf{Models}\end{tabular}}} & GPT-3.5 & 87.0 & 57.2 & 56.7 & 83.1 & 39.3 & 39.2 & 88.6 & 48.8 & - \\ 
\multicolumn{1}{c|}{} & Codex~\cite{EvaluatingLLM} & 91.6 & 67.0 & 55.1 & - & - & - & - & - & - \\ 
\multicolumn{1}{c|}{} & ChatGPT~\cite{Evaluatingchatgpt} & 97.7 & 70.1 & 60.1 & 96.2 & 58.6 & 48.5 & 96.4 & 62.6 & - \\ \cline{2-11} 
\multicolumn{1}{c|}{} & Ours &
 99.0 &
 \textbf{87.8} &
 \textbf{84.8} &
 \textbf{99.0} &
 \textbf{79.4} &
 \textbf{75.9} &
 \textbf{99.4} &
 \textbf{74.2} &
 - \\ \hline 
\end{tabular}}
\label{tab:main}
\end{table*}

\footnotetext[4]{``-" indicates the results are not available. The Codex model could not be reproduced due to an invalid Codex'api. The TS metric did not apply to the Spider-DK dataset.}

We present a comparison between LLM-based models and full-data fine-tuned models which are SOTA as shown in Table \ref{tab:main}.
Ours outperforms all models in almost all evaluation metrics, except for the Spider dataset where the VA is only 0.1 worse than the next best model (RESDSQL-3B). 
LLM-based models may face difficulty generating SQL queries that conform to strict syntactical and semantic rules, resulting in lower VA scores, compared with the fine-tuned models.
We further investigate the effectiveness of ours and compare it to the models using LLMs directly. We observe that the improper utilization of LLMs is generally less effective in generating complex SQL commands. Additionally, the regular LLM-based models are confused with selecting the appropriate schema items required.
On the contrary, ours demonstrates its effectiveness in generating accurate and semantically meaningful SQL queries.
\subsection{Ablation Study}
We investigate the contributions of various components of ours as shown in Table \ref{tab: Components Performance}:
(1) DESEM+P, which uses Poincaré distance for schema-related detection in de-semanticization; 
(2) DESEM-P+E, which uses Euclidean distance for schema-related detection in de-semanticization; 
(3) -DESEM, which retrieves questions without de-semanticization using cosine similarity; 
(4) -Skeleton Retrieval, which demonstrates random examples without retrieval; 
(5) -Schema Filtering, which uses the full range of schema without filtering; 
(6) -SQL Revision, which directly outputs the generated SQL without revision.

\begin{table*}[]
\centering
\renewcommand\arraystretch{1}
\tabcolsep=1.5mm
\small
\caption{Ablation study of different modules. ``-" means not using that strategy, while ``+" means using that strategy.}
\begin{tabular}{lccccccccc}
\hline
\multicolumn{1}{c|}{\multirow{2}{*}{Methods / Datasets}} &
 \multicolumn{3}{c|}{\textbf{SPIDER}} &
 \multicolumn{3}{c|}{\textbf{SPIDER-SYN}} &
 \multicolumn{3}{c}{\textbf{SPIDER-DK}} \\ \cline{2-10} 
\multicolumn{1}{c|}{} &
 \textbf{VA} &
 \textbf{EX} &
 \multicolumn{1}{c|}{\textbf{TS}} &
 \textbf{VA} &
 \textbf{EX} &
 \multicolumn{1}{c|}{\textbf{TS}} &
 \textbf{VA} &
 \textbf{EX} &
 \textbf{TS} \\ \hline
\multicolumn{1}{l|}{DESEM +P (Ours)} &
 \textbf{99.0} &
 \textbf{87.8} &
 \multicolumn{1}{c|}{\textbf{84.8}} &
 \textbf{99.0} &
 \textbf{79.4} &
 \multicolumn{1}{c|}{\textbf{75.9}} &
 \textbf{99.4} &
 \textbf{74.2} &
 \textbf{69.7} \\
 \multicolumn{1}{l|}{DESEM -P+E} & 96.6 & 84.6 & \multicolumn{1}{c|}{79.6} & 97.1 & 77.2 & \multicolumn{1}{c|}{74.3} & 99.1 & 71.3 & 68.2 \\
\multicolumn{1}{l|}{-DESEM} & 93.5 & 71.6 & \multicolumn{1}{c|}{69.8} & 94.7 & 59.2 & \multicolumn{1}{c|}{57.7} & 86.5 & 56.8 & 55.0 \\

\multicolumn{1}{l|}{-Skeleton Retrieval} & 94.3 & 66.9 & \multicolumn{1}{c|}{65.9} & 92.9 & 56.8 & \multicolumn{1}{c|}{55.3} & 93.3 & 59.4 & 57.9 \\

\multicolumn{1}{l|}{-Schema Filtering} & 96.7 & 79.6 & \multicolumn{1}{c|}{78.3} & 97.3 & 73.6 & \multicolumn{1}{c|}{71.5} & 98.5 & 68.6 & 64.3 \\
\multicolumn{1}{l|}{-SQL Revision} & 98.5 & 81.2 & \multicolumn{1}{c|}{78.2} & 95.0 & 69.9 & \multicolumn{1}{c|}{67.3} & 97.6 & 71.8 & 67.1 \\
 \hline 
\end{tabular}

\label{tab: Components Performance}
\end{table*}

Using Poincaré distance instead of Euclidean distance for schema detection (DESEM + P) improves efficacy. Removing de-semanticization (-DESEM) reduces accuracy, highlighting the importance of relevant examples for utilizing SQL components and expressing intent.
Removing skeleton retrieval and comparing question sequences directly lowers performance, indicating domain knowledge's impact on the SPIDER-DK. 
SQL revision enhances generated SQL accuracy, as seen in higher VA scores of methods with revision. 
Schema filtering boosts performance by narrowing the schema range, even without revision.

\subsection{Case Study}
\begin{figure}[H]
\centering
\includegraphics[width=1.0\textwidth]{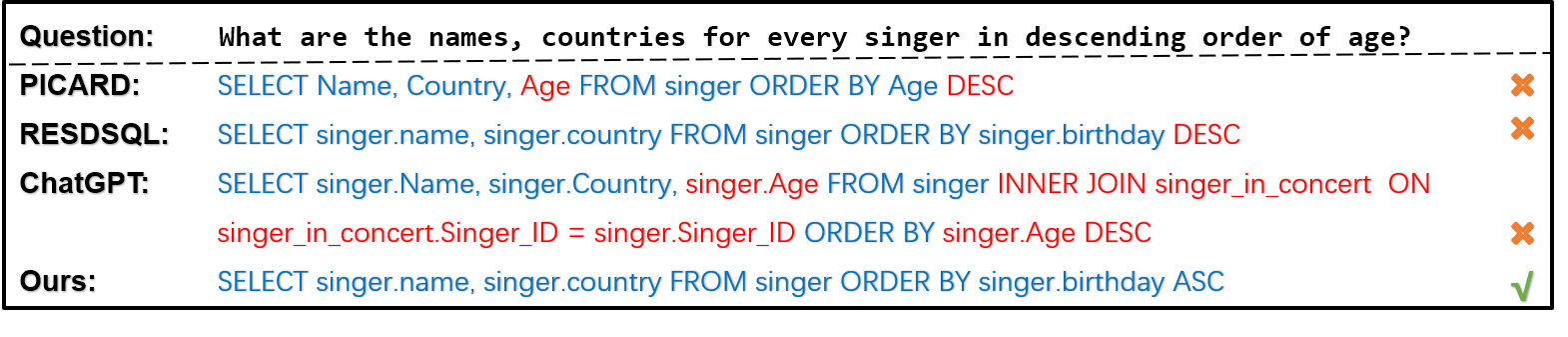}
 \caption{An illustrative case from Spider-DK~\cite{dk}. Blue is the correct generation, and red is the wrong generation.} \label{fig:case}
\end{figure}

To illustrate our method, we show a comparison of predicted SQLs in Fig.~\ref{fig:case} using PICARD, RESDSQL, ChatGPT, and our approach.

\vspace{-5mm}

It shows that PICARD and ChatGPT generate an extra column ``Age", and the three models fall short of accurately capturing the desired sorting order.
This is because the database schema only has a ``birthday" column, and sorting by age is equivalent to sorting by ``birthday" in ascending order. ChatGPT attempts to generate a more comprehensive SQL query by including an unnecessary JOIN operation with the singer\_in\_concert table.
Ours considers the requirement of ordering the results, albeit in the opposite direction specified in the question. However, fine-tuned models like PICARD and RESDSQL may struggle with complex questions due to limitations in learned patterns and structures.

\section{Conclusion and Future Work}
We propose a Text-to-SQL generation framework that prompts LLMs with few retrieved demonstrations. A limitation of ours is over-reliance on LLMs's ability for SQL code generation. LLMs with certain capabilities can work well with our method. Future research will focus on external knowledge reasoning, as well as efficiency when dealing with large databases. Our approach can be generalized to knowledge base question answering and code generation tasks.


\bibliographystyle{splncs03_unsrt}
\bibliography{egbib.bib}

\begin{thebibliography}{10}
\providecommand{\url}[1]{\texttt{#1}}
\providecommand{\urlprefix}{URL }

\bibitem{ratsql}
Wang, B., Shin, R., Liu, X., Polozov, O., Richardson, M.: Rat-sql:
  Relation-aware schema encoding and linking for text-to-sql parsers. ACL
  (2020)

\bibitem{Cai_Xu_Zhang_Yang_Li_Liang_2018_Encoder-Decoder}
Cai, R., Xu, B., Zhang, Z., Yang, X., Li, Z., Liang, Z.: An encoder-decoder
  framework translating natural language to database queries. In: IJCAI (Jul
  2018)

\bibitem{li2023resdsql}
Li, H., Zhang, J., Li, C., Chen, H.: Decoupling the skeleton parsing and schema
  linking for text-to-sql. arXiv:2302.05965  (2023)

\bibitem{Graphix-T5}
Li, J., Hui, B., et~al.: Graphix-t5: Mixing pre-trained transformers with
  graph-aware layers for text-to-sql parsing. arXiv:2301.07507  (2023)

\bibitem{zhao2023survey}
Zhao, W.X., Zhou, K., Li, J., et~al.: A survey of large language models. arXiv
  preprint arXiv:2303.18223  (2023)

\bibitem{GPT3}
Brown, T., Mann, B., Ryder, N., et~al.: Language models are few-shot learners.
  NIPS  33,  1877--1901 (2020)

\bibitem{Codex}
Chen, M., Tworek, J., Jun, H., Yuan, Q., Pinto, H.P.d.O., et~al.: Evaluating
  large language models trained on code. arXiv:2107.03374  (2021)

\bibitem{bert}
Devlin, J., Chang, M.W., Lee, K., Toutanova, K.: Bert: Pre-training of deep
  bidirectional transformers for language understanding. NAACL  (2018)

\bibitem{RoBERTa}
Zhuang, L., Wayne, L., Ya, S., Jun, Z.: A robustly optimized bert pre-training
  approach with post-training. In: CCL. pp. 1218--1227 (2021)

\bibitem{BART}
Lewis, M., Liu, Y., et~al.: Bart: Denoising sequence-to-sequence pre-training
  for natural language generation, translation, and comprehension. In: ACL (Jul
  2020)

\bibitem{t5}
Raffel, C., Shazeer, N., et~al.: Exploring the limits of transfer learning with
  a unified text-to-text transformer. JMLR  21,  5485--5551 (2020)

\bibitem{EvaluatingLLM}
Rajkumar, N., Li, R., Bahdanau, D.: Evaluating the text-to-sql capabilities of
  large language models. arXiv:2204.00498  (2022)

\bibitem{Evaluatingchatgpt}
Liu, A., Hu, X., Wen, L., Yu, P.S.: A comprehensive evaluation of chatgpt's
  zero-shot text-to-sql capability. arXiv:2303.13547  (2023)

\bibitem{Binding}
Cheng, Z., Xie, T., Shi, P., et~al.: Binding language models in symbolic
  languages. In: ICLR (2023)

\bibitem{picard}
Scholak, T., Schucher, N., Bahdanau, D.: Picard: Parsing incrementally for
  constrained auto-regressive decoding from language models. EMNLP  (2021)

\bibitem{rasat}
Qi, J., Tang, J., He, Z., et~al.: {RASAT:} integrating relational structures
  into pretrained seq2seq model for text-to-sql. In: EMNLP. pp. 3215--3229
  (2022)

\bibitem{lee2022does}
Lee, Y.J., Lim, C.G., Choi, H.J.: Does gpt-3 generate empathetic dialogues? a
  novel in-context example selection method and automatic evaluation metric for
  empathetic dialogue generation. In: COLING. pp. 669--683 (2022)

\bibitem{Su_Kasai_2022}
Su, H., Kasai, J., et~al.: Selective annotation makes language models better
  few-shot learners. arXiv:2209.01975  (2022)

\bibitem{Rubin_Herzig_Berant}
Rubin, O., Herzig, J., Berant, J.: Learning to retrieve prompts for in-context
  learning. In: NAACL. pp. 2655--2671 (2022)

\bibitem{irnet}
Guo, J., et~al.: Towards complex text-to-sql in cross-domain database with
  intermediate representation. In: ACL. pp. 4524--4535 (2019)

\bibitem{Bogin_Berant_Gardner_2019}
Bogin, B., Berant, J., Gardner, M.: Representing schema structure with graph
  neural networks for text-to-sql parsing. In: ACL (Sep 2019)

\bibitem{Chen_Chen_Zhao_Cao_Xu_Zhu_Yu_2021_ShadowGNN}
Chen, Z., Chen, L., Zhao, Y., Cao, R., Xu, Z., Zhu, S., Yu, K.: Shadowgnn:
  Graph projection neural network for text-to-sql parser. In: NAACL (Jun 2021)

\bibitem{LGESQL}
Cao, R., Chen, L., et~al.: Lgesql: Line graph enhanced text-to-sql model with
  mixed local and non-local relations. In: ACL (Jul 2021)

\bibitem{Hui_Geng_Ren_Li_Li_Sun_Huang_Si_Zhu_Zhu_2021}
Hui, B., Geng, R., Ren, Q., et~al.: Dynamic hybrid relation exploration network
  for cross-domain context-dependent semantic parsing. AAAI  (May 2021)

\bibitem{hui-etal-2022-s2sql}
Hui, B., Geng, R., Wang, L., et~al.: S2sql: Injecting syntax to question-schema
  interaction graph encoder for text-to-sql parsers. In: ACL. pp. 1254--1262
  (2022)

\bibitem{STAR_Cai_2022}
Cai, Z., Li, X., Hui, B., Yang, M., Li, B., et~al.: Star: Sql guided
  pre-training for context-dependent text-to-sql parsing. EMNLP  (Oct 2022)

\bibitem{lin2020bridging}
Lin, X.V., Socher, R., Xiong, C.: Bridging textual and tabular data for
  cross-domain text-to-sql semantic parsing. In: EMNLP. pp. 4870--4888 (2020)

\bibitem{X-SQL}
He, P., Mao, Y., Chakrabarti, K., Chen, W.: X-sql: reinforce schema
  representation with context. arXiv:1908.08113  (2019)

\bibitem{hybrid}
Lyu, Q., Chakrabarti, K., Hathi, S., Kundu, S., Zhang, J., Chen, Z.: Hybrid
  ranking network for text-to-sql. arXiv preprint arXiv:2008.04759  (2020)

\bibitem{GAZP}
Zhong, V., Lewis, M., Wang, S.I., Zettlemoyer, L.: Grounded adaptation for
  zero-shot executable semantic parsing. In: EMNLP. pp. 6869--6882 (2020)

\bibitem{choi-etal-2021-ryansql}
Choi, D., Shin, M.C., Kim, E., Shin, D.R.: Ryansql: Recursively applying
  sketch-based slot fillings for complex text-to-sql in cross-domain databases.
  CL  (2021)

\bibitem{yu2021similar}
Yu, W., et~al.: Similar questions correspond to similar sql queries: A
  case-based reasoning approach for text-to-sql translation. In: ICCBR. pp.
  294--308. Springer (2021)

\bibitem{tian2019learning}
Tian, Z., Bi, W., Li, X., Zhang, N.L.: Learning to abstract for
  memory-augmented conversational response generation. In: ACL. pp. 3816--3825
  (2019)

\bibitem{song2022retrieval}
Song, Y., et~al.: Retrieval bias aware ensemble model for conditional sentence
  generation. In: ICASSP. pp. 6602--6606. IEEE (2022)

\bibitem{wen2023grace}
Wen, Z., et~al.: Grace: Gradient-guided controllable retrieval for augmenting
  attribute-based text generation. In: Findings of ACL 2023. pp. 8377--8398
  (2023)

\bibitem{Hyperbolic}
Ganea, O., B{\'e}cigneul, G., Hofmann, T.: Hyperbolic neural networks. NIPS
  (2018)

\bibitem{Probing_BERT}
Chen, B., et~al.: Probing bert in hyperbolic spaces. arXiv:2104.03869  (2021)

\bibitem{spider}
Yu, T., Zhang, R., et~al.: Spider: A large-scale human-labeled dataset for
  complex and cross-domain semantic parsing and text-to-sql task. In: EMNLP
  (Jun 2019)

\bibitem{syn}
Gan, Y., Chen, X., Huang, Q., Purver, M., et~al: Towards robustness of
  text-to-sql models against synonym substitution. In: ACL (Jul 2021)

\bibitem{dk}
Gan, Y., Chen, X., Purver, M.: Exploring underexplored limitations of
  cross-domain text-to-sql generalization. In: EMNLP (Dec 2021)

\bibitem{ts}
Zhong, R., Yu, T., Klein, D.: Semantic evaluation for text-to-sql with
  distilled test suites. In: EMNLP. pp. 396--411 (2020)

\bibitem{FAISS}
Johnson, J., Douze, M., Jegou, H.: Billion-scale similarity search with gpus.
  IEEE Transactions on Big Data p. 535–547 (Jun 2019)

\end{thebibliography}
\end{document}